# Developing a Machine Learning Algorithm-Based Classification Models for the Detection of High-Energy Gamma Particles

Emmanuel A. Dadzie, *and* Kelvin K. Kwakye

*Abstract*— Cherenkov gamma telescope observes high energy gamma rays, taking advantage of the radiation emitted by charged particles produced inside the electromagnetic showers initiated by the gammas, and developing in the atmosphere. The detector records and allows for the reconstruction of the shower parameters. The reconstruction of the parameter values was achieved using a Monte Carlo simulation algorithm called CORSIKA. The present study developed multiple machine-learning-based classification models and evaluated their performance. Different data transformation and feature extraction techniques were applied to the dataset to assess the impact on two separate performance metrics. The results of the proposed application reveal that the different data transformations did not significantly impact (p = 0.3165) the performance of the models. A pairwise comparison indicates that the performance from each transformed data was not significantly different from the performance of the raw data. Additionally, the SVM algorithm produced the highest performance score on the standardized dataset. In conclusion, this study suggests that high-energy gamma particles can be predicted with sufficient accuracy using SVM on a standardized dataset than the other algorithms with the various data transformations.

## I. INTRODUCTION

THE data are Monte Carlo generated using an algorithm described in [1] to simulate registration of high energy gamma particles in a ground-based atmospheric Cherenkov gamma telescope using the imaging technique. Cherenkov gamma telescope observes high energy gamma rays, taking advantage of the radiation emitted by charged particles produced inside the electromagnetic showers initiated by the gammas, and developing in the atmosphere. This Cherenkov radiation (of visible to UV wavelengths) leaks through the atmosphere and gets recorded in the detector, allowing reconstruction of the shower parameters. The available information consists of pulses left by the incoming Cherenkov photons on the photomultiplier tubes, arranged in a plane in the camera. Depending on the energy of the primary gamma, a total of a few hundred to some 10000 Cherenkov photons get collected, in patterns (called the shower image), allowing to discriminate statistically those caused by primary gammas (signal) from the images of hadronic showers initiated by cosmic rays in the upper atmosphere (background) for defect detection in concrete, especially with limited data.

Typically, the image of a shower after some pre-processing is an elongated cluster. Its long axis is oriented towards the camera center if the shower axis is parallel to the telescope's optical axis, i.e., if the telescope axis is directed towards a point source. The principal component analysis is performed in the camera plane, which results in a correlation axis and defines an ellipse. If the depositions were distributed as a bivariate Gaussian, this would be an equidensity ellipse. The characteristic parameters of this ellipse (often called Hillas parameters) are among the image parameters that can be used for discrimination. The energy depositions are typically asymmetric along the major axis, and this asymmetry can also be used in discrimination. There are, in addition, further discriminating characteristics, like the extent of the cluster in the image plane, or the total sum of depositions.

## II. RELATED WORK

Much research has employed different modeling techniques to create classification models for the separation of gamma particles from hadron (background). Some researchers investigated the performance of multiple-layer perceptron (MLP) for supervised classification, and self-organizing tree algorithm (SOTA) for unsupervised classification in the separation of gamma rays from the background. These authors further combined these techniques and suggested that their approach significantly reduced training time and produced better classification results [2]. Whereas others use unity embeddings [3-5] to improve classification results.

Other researchers used a tree classification method, Random Forest (RF), to analyze the data of a ground-based gamma telescope. The RF method showed superior performance when compared with other traditional semi-empirical techniques. The further discussed critical issues of the method and elaborated on the use of RF in the estimation of continuous parameters from other variables [6]. Some authors applied a kernel-based fixed-size least square support vector machine (FS-LSSVM) on large-scale datasets, which included the magic gamma dataset. Their

E. Dadzie and K. Kwakye are Human Factors Ph.D. candidates at the Industrial and Systems Engineering Department of the North Carolina Agricultural and Technical State University, 1601 E Market St, Greensboro, NC 27401, USA (email: eadadzie@aggies.ncat.edu).

approach exhibited substantial speedup over the existing FS-LSSVM implementations [7].

III. METHOD

*A. Proposed Approach*

The present study aims to create classification models for the detection of high-energy gamma particles from celestial space. The data is Monte Carlo generated and sourced from the UCI data repository. This study aims at assessing the performance of different classification models developed using various machine learning algorithms. This study further explores different data transformation techniques and the impact on the performance of these models. The details of the proposed approach are elaborated in the following paragraphs.

First, the data is acquired from the repository and studied, which investigated the principles governing the instrument, the parameters, and attributes, and the final measures from the setup. The data is examined for possible missing values representations and the range of each attribute. The next stage in our practice is preprocessing of the data. The missing values and outliers are identified and eliminated from the data. This stage is followed by different transformations of the data to be used in model development. Fig 1 presents the structure of the proposed methodology.

The transformation of the data involves both selecting the relevant features or attributes as one method, and the complete modification of the data points as the other method. The transformation either maintains the dimension of the dataset or reduces it. After the transformation stage, various classification models are created using different machine learning algorithms. Each algorithm is used on all the data transforms separately. Each algorithm is evaluated using two metrics, cross-validation and the area under the curve (AUC).

Finally, the performance of the algorithms on the different data transformations is assessed. The statistical significance between the various data transforms, as well as the raw data is tested based on the cross-validation test and AUC separately.

The primary focus of the current study is to investigate the impact of different data transformation techniques on the performance of machine-learning-based classification models for the detection of high-energy gamma particles from space using the gamma telescope.

*B. Dataset and Preprocessing*

The data is generated using CORSIKA, a Monte Carlo program (code) [1], to simulate the registration of high-energy gamma particles in a ground-based atmospheric Cherenkov gamma telescope using the imaging technique. The CORSIKA algorithm has been used by [8] – [10] in their study in generating data for validation of their experiments. This telescope observes high-energy gamma rays using the radiation emitted by charged particles produced inside the electromagnetic showers initiated by the gammas, and developing in the atmosphere. The dataset has 10 attributes, which are continuous, and a binary class that indicates an instance to be gamma (signal) or hadron (background). The dataset has 19020 instances with no missing values. The dataset

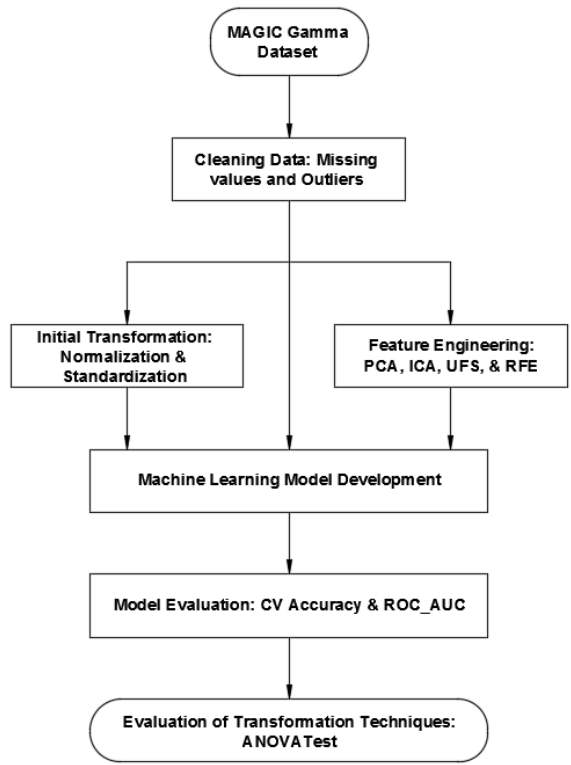

Fig. 2. Structure of the proposed methodology of the study showing all the stages in the process.

is unbalanced with 12332 and 6688 instances for gamma and hadron, respectively.

The preprocessing of the data includes the identification and removal of missing values, which existed the data as black cells, and the numerical value of 99999. The data void of missing values serves as the raw baseline data, which is later compared with the preprocessed and transformed data in the performance evaluation during the model development. Further, outliers are removed from the data using the equations below.

$$Outlier > Q3 + 1.5 * (Q3 - Q1) \quad or \quad Outlier < Q1 - 1.5 * (Q3 - Q1) \quad (1)$$

$$Outlier - mean > |3 * s| \quad (2)$$

Where Q1, Q3, and s represent the first quantile, the third quantile, and the standard deviation of each attribute. Equation 1 identified 975 outliers, whereas Eq 2 identified 2998. Equation 2 identified 3 times the number of outliers identified by Eq 1. Hence, Eq 2 was used to clean the data. This data is considered as the "clean" data in the study from which it was later normalized and standardized for the model development.

*C. Feature Engineering*

This section of the study involves both the selection of relevant features from the group of attributes and the extraction of hidden features within the attributes that account for over 95% of the variance in the data. The selection of relevant features includes univariate feature selection (UFS) and recursive feature elimination (RFE). The extraction of hidden features is achieved using principal component analysis (PCA) and independent component analysis (ICA).

*1) Principal Component Analysis (PCA)*

PCA constructs relevant features by linearly transforming correlated variables into a smaller number of uncorrelated variables, also known as principal components (Joliffe 2002). The resulting principal components are linear combinations of the original data capturing most of the variance in the data (Joliffe 2002; Mourao-Miranda et al. 2005; MouraoMiranda et al. 2006; Zhu et al. 2005). The current study reduced the dimension of the "clean" data to a size that captures at least 95% of the variance in the data.

*2) Independent Component Analysis (ICA)*

ICA is a multivariate data-driven technique that belongs to the broader category of blind-source separation methods used to separate data into underlying independent information components (Stone, 2004). ICA separates a set of mixed features into a set of independent and relevant features. To achieve this goal, ICA assumes the relevant features are statistically independent of an unknown but linear mixing process (Calhoun et al. 2009).

*3) Univariate Feature Selection (UFS)*

This technique selects the best features from the data based on a univariate test. Some of the tests are chi-squared and F-test for classification problems. The chi-squared does not work for attributes with negative values. Hence, an f-test is used in this study. The methods based on F-test estimate the degree of linear dependence between two random variables.

*4) Recursive Feature Elimination (RFE)*

This approach selects features by recursively considering smaller and smaller sets of features. First, the estimator is trained on the initial set of features, and the importance of each feature is obtained by a metric, such as a feature importance. The least important features are pruned from the current set of features. That procedure is recursively repeated on the pruned set until the desired number of features to select is eventually reached.

*D. Classification Model Development*

The 8 different forms of the data, namely, the raw, clean, normalized, standardized, PCA, ICA, UFS, and RFE data is used to create different classification models using six separate machine learning algorithms. The algorithms selected for the models are logistic regression (LR), linear discriminant analysis (LDA), k-nearest neighbor (KNN), decision tree (CART), naïve Bayes (NB), and support vector machine (SVM). The models were evaluated for performance using two different metrics, cross-validation accuracy and the area under the curve (AUC) of the receiver operating characteristics (ROC). Each data form produced 6 classification models for each performance metric. The results used further used to check for statistical significance between the different data transformation approaches in creating a classification for the detection of high-energy gamma particles.

*E. Data Analysis*

To prove if the different data transformation techniques had an impact on the classification performance of the model, an analysis of variance (ANOVA) is performed. The independent variable (IV) in this case is the different transformations including the raw data. The dependent variables (DVs) are accuracy and ROC_AUC but were analyzed separately. A pairwise comparison test is conducted to identify the pair of transformation approaches that are significantly different or not in terms of classification performance. Furthermore, the performance of each transformation is compared to the performance of the raw data to evaluate the statistical significance. The ANOVA test was preceded by a test of adequacy to implement this kind of analysis. The adequacy checks include normality, homogeneity, and equality of variance test on the dataset, that is, the CV accuracy or the ROC_AUC.

*F. Software Packages*

The current study utilized a few software packages to achieve its set goals, depending on its functionalities. Statistical analysis of data was performed using SAS statistical packages version 9.4. Secondly, the open-source python program version 3.7.1, with its numerous packages and modules facilitated the development of machine learning algorithms and data visualization for the study. The Scikit Learn package has inbuilt modules for most of the machine learning algorithms. Other packages such as Matplotlib and Seaborn enabled the visualization of data and results from the various algorithms.

TABLE I
SKEWNESS VALUES FOR THE ATTRIBUTES FOR THE DATASET.

| Attribute | Initial values | Final values |
|---|---|---|
| fLength | 137.187880 | 1.146622 |
| fWidth | 97.449352 | 0.811858 |
| fSize | 137.902051 | 0.695432 |
| fConc | 97.506405 | 0.445778 |
| fConc1 | 79.605274 | 0.543348 |
| fAsym | 136.533855 | -0.313774 |
| fM3Long | 38.171009 | 0.138903 |
| fM3Trans | 52.071776 | -0.005655 |
| fAlpha | 45.900631 | 0.937793 |
| fDist | 79.180620 | 0.139228 |

The Pandas package enabled the organization and manipulation of data structures.

IV. RESULTS

The current application begins with the preparation of the data to remove unwanted data points that could introduce biases in the classification model. The data preparation is also aimed at maximizing the performance of the models.

Data Preprocessing

*1) Removing missing values and outliers*

The cleaning of the data starts with the removal of missing values that existed in the data as empty cells and a numerical value of 99999. A total of 123 missing values were identified and removed. The next process was the removal of outliers, which were identified by using (1) and (2). Equation 1 removed 975 outliers, whereas (2) removed 2998 outliers. Equation (2) was employed as the approach for outlier removal in this study. The skewness of each attribute before and after cleaning the data is presented in Table I, and the scatter matrix plotted before and after cleaning the data is displayed in Fig 2.

The skewness values show that each attribute was highly skewed before the data was cleaned. The high values are due to the missing values that were represented by the numerical value of 99999. The removal of those values together with the outlier made the attributes near normal, with the skewness values approaching zero. Additionally, the visualization presented by the scatter matrix shows that the dataset with the unwanted values produced plots with unrecognizable patterns. However, when the data was cleaned, the plots produced distinctive patterns for each pair of attributes, as well as the near-normal histograms for each attribute.

*2) Initial data transformation*

The initial transformation of the data did not result in a reduction of features or dimensions. This transformation was employed to remove the biases from the data caused by a different range of values for the different attributes. Two transformations were performed. The first was normalization, which converted most of the values into a range between 0 and 1. The second transformation converted the values in each attribute into a z score with most of the values between -3 and +3.

*3) Feature selections and extraction*

This stage is aimed at removing redundancies from the data by either transforming the data to capture most of the variance while reducing the dimensions or simply selecting the relevant attributes that contribute to the majority of the variance.

*Results for (PCA)*

This feature extraction technique captured the hidden feature within the dataset by ensuring that at least 95% of the variance is accounted for in the dataset. Below are the first principal component and the explained variance ratios for the 5 components that accounted for 97.69% of the variance.

**PC1** = 1.849e-01 * **fLength** + 4.177e-02 * **fWidth** + 2.561e-03 * **fSize** - 8.551e-04 * **fConc** - 4.759e-04 * **fConc1** - 4.753e-02 * **fAsym** + 1.449e-01 * **fM3Long** - 1.964e-04 * **fM3Trans** - 1.333e-01 * **fAlpha** + 9.607e-01 * **fDist**

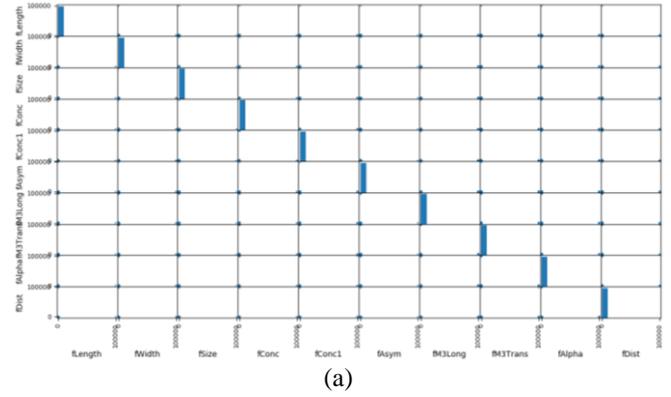

(a)

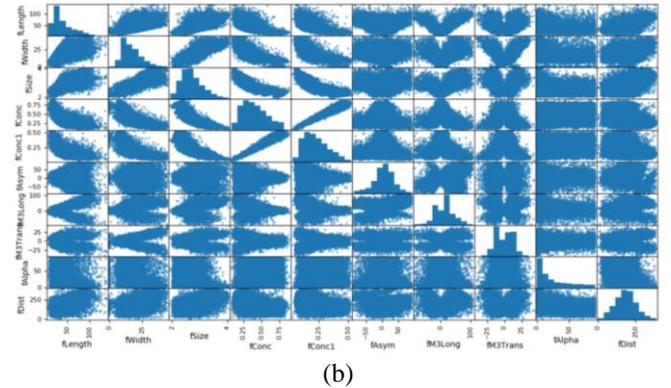

(b)

Fig. 2. The scatter matrix for the dataset before (a) and after (b) the removal of missing values and outliers. The diagonal of the matrix shows the histogram plot for each attribute.

PCA Ratios of explained variance: [0.627236, 0.165211, 0.082783, 0.067893, 0.033773]

*Results for ICA*

This technique also extracted five independent components after performing 200 iterations. The first component is present below.

**PC1** = -3.189e-04 * **fLength** - 7.470e-05 * **fWidth** - 3.692e-06 * **fSize** + 1.838e-06 * **fConc** + 1.059e-06 * **fConc1** - 1.533e-06 * **fAsym** + 2.569e-04 * **fM3Long** + 7.453e-04 * **fM3Trans** - 3.572e-05 * **fAlpha** - 5.010e-07 * **fDist**

*Results for UFS*

The same number of features was selected for this technique. The feature score and its ranking are as presented below. Only the top five ranked attributes were selected for the model development.

Feature scores: [7.577e+00 **3.009e+02** 8.855e+01 **3.225e+02 3.807e+02** 2.206e+01 **1.540e+02** 8.226e-01 **3.882e+03** 3.068e+01]

Feature ranking: [2, **7**, 5, **8**, **9**, 3, **6**, 1, **10**, 4]

The selected attributes are fAlpha, fConc1, fConc, fWidth, and fM3Long, respectively.

*Results for RFE*
This technique was also implemented to select the top 5 features from the dataset. Below are the details of the output of this technique. A numerical value of 1 in the feature ranking indicates that the feature is selected.

Selected Features: [False **True True True True** False False False *True* False]

Feature Ranking: [2 **1 1 1 1** 4 3 6 **1** 5]

The selected attributes correspond to fWidth, fSize, fConc, fConc1, and fAlpha, respectively.

B. *Machine learning classification models*
1) *The outcome of the cross-validation (CV) test*
The results show some level of consistency within the values, but some of the transformations produced very low mean accuracy mainly in the SVM model. Four out of the seven transformations produced higher mean accuracy than the raw (baseline) dataset. The transformation that produced the highest accuracy is the standardization of the data. Table II presents the summary results for the CV-evaluated model output.

2) *Outcome of ROC_AUC*
The results indicate relatively consistent values with similar variability across the different transformations. Only 2 out of the 7 transformation techniques produced a higher mean AUC than the raw (baseline) dataset. The standardization transformation produced the highest mean AUC. Table III displays the summary results for the AUC-evaluated model output.

C. *Evaluation of transformation techniques on performance*
1) *ANOVA on cross-validation accuracy*
The model adequacy checks revealed that the dataset derived from the results of the CV test was not normal and homogeneous. Hence, the ANOVA test could not be performed to evaluate the statistical significance between the different transforms. However, the results indicate that the standardized data produced the highest mean accuracy.

2) *ANOVA on ROC_AUC*
The adequacy test reveals that the dataset derived from the AUC results in Table III was normal, homogeneous, and had similar variance using a significance level of 0.05. Fig 3 shows the distribution of the residuals of the data that depicts a normal distribution.
The ANOVA test shows that at 0.05 significance level, the AUC values produced by the various transformations were not significantly different (p = 0.3165) from each other based on the evidence given by the dataset. Hence, the various transformations did not have any impact on the performance of the classification models. Fig 4 shows the SAS output of the

TABLE II
SUMMARY RESULTS OF THE PERFORMANCE OF MODELS BASED ON CROSS-VALIDATION ACCURACY.

| Data | LR | LDA | KNN | CART | NB | SVM | Mean |
|---|---|---|---|---|---|---|---|
| Raw | 0.7604 | 0.7606 | 0.7531 | 0.7944 | 0.6435 | 0.5148 | 0.7045 |
| Clean | 0.7379 | 0.7385 | 0.7094 | 0.7440 | 0.6773 | 0.5019 | 0.6849 |
| Norm | 0.7377 | 0.7385 | 0.7646 | 0.7450 | 0.6773 | 0.7533 | *0.7361 |
| Stand | 0.7408 | 0.7385 | 0.7648 | 0.7452 | 0.6773 | 0.8215 | ***0.7480** |
| PCA | 0.7314 | 0.7262 | 0.7112 | 0.6712 | 0.7181 | 0.5077 | 0.6777 |
| ICA | 0.7198 | 0.7262 | 0.7160 | 0.6944 | 0.7115 | 0.4860 | 0.6757 |
| UFS | 0.7152 | 0.7135 | 0.7258 | 0.7044 | 0.6883 | 0.7002 | *0.7079 |
| RFE | 0.7285 | 0.7277 | 0.7050 | 0.7273 | 0.6942 | 0.7383 | *0.7202 |

\* Indicate mean performance values that are higher than the value for raw (baseline) data. The bold value is the highest value.

TABLE III
SUMMARY RESULTS OF THE PERFORMANCE OF MODELS BASED ON ROC_AUC.

| Data | LR | LDA | KNN | CART | NB | SVM | Mean |
|---|---|---|---|---|---|---|---|
| Raw | 0.8394 | 0.8364 | 0.8264 | 0.7901 | 0.7558 | 0.6979 | 0.7910 |
| Clean | 0.7958 | 0.7975 | 0.7768 | 0.7502 | 0.7342 | 0.6672 | 0.7536 |
| Norm | 0.7976 | 0.7975 | 0.8441 | 0.7460 | 0.7342 | 0.8312 | *0.7918 |
| Stand | 0.7967 | 0.7975 | 0.8410 | 0.7473 | 0.7342 | 0.8964 | ***0.8022** |
| PCA | 0.7780 | 0.7789 | 0.7793 | 0.6711 | 0.7816 | 0.6661 | 0.7425 |
| ICA | 0.7757 | 0.7789 | 0.7892 | 0.6944 | 0.7813 | 0.7759 | 0.7659 |
| UFS | 0.7813 | 0.7816 | 0.7887 | 0.7085 | 0.7486 | 0.7389 | 0.7579 |
| RFE | 0.7863 | 0.7865 | 0.7653 | 0.7258 | 0.7451 | 0.8079 | 0.7695 |

\* Indicate mean performance values that are higher than the value for raw (baseline) data. The bold value is the highest.

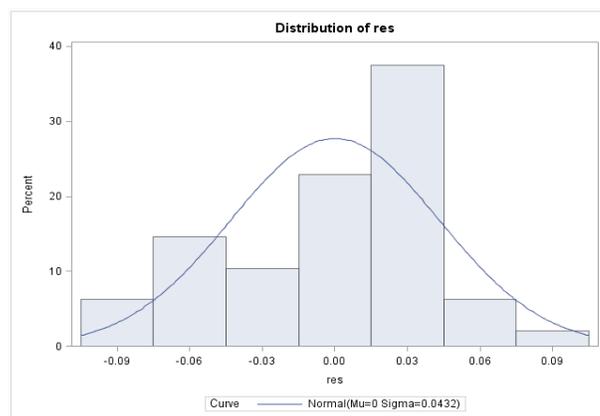

Fig. 3. The distribution of the residuals derived from the AUC summary results.

ANOVA test.

Fig. 4. The output of the ANOVA test from the SAS statistical software.

Furthermore, Dunnett's pairwise comparison test was performed to assess the significance in performance between each transformation and the raw (baseline) data. The SAS output of this test is presented in Fig 5. The results indicate that at 0.05 significance level, the performance produced by the raw data was not significantly different from the performance produced by all the other transformed data.

Fig. 5. The output of Dunnett's pairwise comparison test from the SAS statistical software.

However, the results from Table III show that the standardized data produced the highest mean AUC despite the difference not being statistically significant. This output of the results for the AUC is visualized in Fig 6.

## V. DISCUSSION AND CONCLUSIONS

The present study developed multiple classification models for the detection of high-energy gamma particles using different machine learning algorithms. The proposed methods were to maximize the classification performance of the algorithms by using different transformation techniques. These techniques included initial transformation, which maintained the dimension of the data, and the other resulted in dimensionality reduction. Each algorithm was trained and tested on all the data forms, that is, both raw and transformed data. Two performance metrics, cross-validation accuracy, and ROC_AUC were used to evaluate the performance of the models. The effect of the data transformations was finally assessed using the ANOVA test.

The results indicated similar performance levels for all the models across the different datasets (i.e., raw, clean, normalized, standardized, PCA, ICA, UFS, and RFE transformed data). The ANOVA test reveals that the performance levels across the different transformations were not significantly different. The pairwise comparison of the performance (i.e., AUC) between the raw data and all the other data forms were not significantly different. Hence, none of the transformations increase the performance significantly. However, the mean accuracy for normalized, standardized, UFS, and RFE transformations were higher than that of the raw (baseline) data. Similarly, in the case of AUC, only the normalized and standardized were higher than the value for the raw data. In both cases of the performance metrics, the standardization transformation produced the highest score.

In conclusion, the present study developed multiple classification models using different machine learning algorithms and different data transformations. The results suggest that models created with the raw data have similar classification performance as the models created with the transformed data. However, the best model for the detection of high-energy gamma particles is the support vector machine (SVM) algorithm on a standardized dataset.

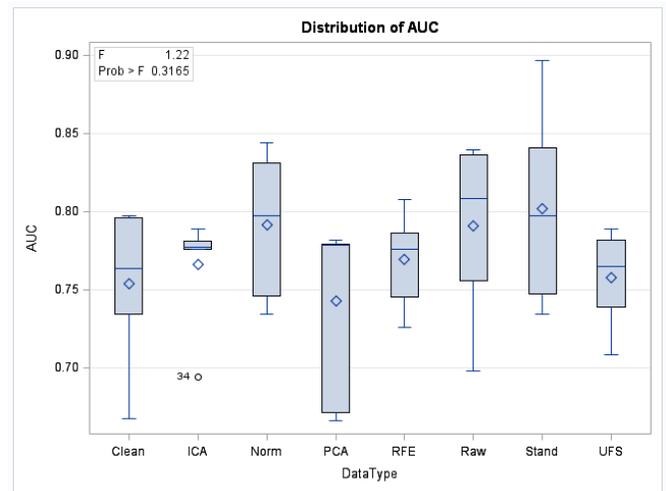

Fig. 6. Box and whisker plot of the results from the AUC performance based on the different data transformation techniques.

## REFERENCES

[1] Heck, D., Schatz, G., Knapp, J., Thouw, T., & Capdevielle, J. N. (1998). CORSIKA: a Monte Carlo code to simulate extensive air showers (No. FZKA-6019).

[2] Boinee, P., Barbarino, F., De Angelis, A., Saggion, A., & Zacchello, M. (2006). Neural networks for gamma-hadron separation in MAGIC. In Frontiers of Fundamental Physics (pp. 297-302). Springer, Dordrecht.

[3] Aboah, A., & Adu-Gyamfi, Y. (2020). Smartphone-Based Pavement Roughness Estimation Using Deep Learning with Entity Embedding. *Advances in Data Science and Adaptive Analysis*, *12*(03n04), 2050007.

[4] Aboah, A., Boeding, M., & Adu-Gyamfi, Y. (2021). Mobile Sensing for Multipurpose Applications in Transportation. *arXiv preprint arXiv:2106.10733*.

[5] Kwakye, K., Seong, Y., & Yi, S. (2020, August). An Android-based mobile paratransit application for vulnerable road users. In *Proceedings of the 24th Symposium on International Database Engineering & Applications* (pp. 1-5).

[6] Albert, J., Aliu, E., Anderhub, H., Antoranz, P., Armada, A., Asensio, M., ... & Becker, J. (2008). Implementation of the random forest method for the imaging atmospheric Cherenkov telescope MAGIC. Nuclear Instruments and Methods in Physics Research Section A: Accelerators, Spectrometers, Detectors, and Associated Equipment, 588(3), 424-432.

[7] Chandorkar, M., Mall, R., Lauwers, O., Suykens, J. A., & De Moor, B. (2015, December). Fixed-size least squares support vector machines: Scala implementation for large-scale classification. In 2015 IEEE Symposium Series on Computational Intelligence (pp. 522-528). IEEE.

[8] Bock, R. K., Chilingarian, A., Gaug, M., Hakl, F., Hengstebeck, T., Jiřina, M., ... & Vaiciulis, A. (2004). Methods for multidimensional event classification: a case study using images from a Cherenkov gamma-ray telescope. Nuclear Instruments and Methods in Physics Research Section A: Accelerators, Spectrometers, Detectors, and Associated Equipment, 516(2-3), 511-528.

[9] Savický, P., & Kotrc, E. (2004). Experimental study of leaf confidences for random forest. In Proceedings of the 16th Symposium on Computational Statistics (pp. 1767-1774). Prague, Czech Republic.

[10] Dvořák, J., & Savický, P. (2007, April). Softening splits in decision trees using simulated annealing. In International Conference on Adaptive and Natural Computing Algorithms (pp. 721-729). Springer, Berlin, Heidelberg.